\documentclass{article}
\usepackage{iclr2027_conference,times}
\usepackage[T1]{fontenc}
\usepackage[utf8]{inputenc}
\usepackage{xcolor,hyperref,amsmath,amssymb,graphicx,booktabs,tabularx,array}
\usepackage{flafter}
\hypersetup{colorlinks=true,linkcolor=black,citecolor=black,urlcolor=black,pdfauthor={Mainak Mallick and Seung-Kyum Choi},pdftitle={When Labels Are Scarce: An Oscillatory State Space Model for Vibration Diagnosis}}
\title{When Labels Are Scarce: An Oscillatory State Space Model for Vibration Diagnosis}
\author{Mainak Mallick \quad Seung-Kyum Choi\\Georgia Institute of Technology}
\iclrfinalcopy
\begin{document}\maketitle

\begin{abstract}
Machine fault diagnosis from vibration requires learning from scarce labelled
fault recordings while meeting the computational constraints of edge devices
for local inference.
We introduce DualRes, a compact oscillatory state-space model that combines
two complementary spectral views of vibration, capturing rapid changes and
fine frequency structure. Time-aligned views are processed by selective
oscillatory memory, which learns how long to retain temporal patterns.
The encoder contains 39,528 parameters. We evaluate supervised learning across six bearing datasets and
a gearbox benchmark, with an additional gearbox pilot. Recording-level splits
and explicit accounting of labelled duration distinguish data efficiency from
repeated exposure to correlated samples. On the main gearbox benchmark,
DualRes achieves state-of-the-art performance among the nine evaluated methods
at six of seven label budgets. With about six labelled seconds per class, it
improves macro-F1 by 16.1 percentage points over the next strongest comparator. On the same benchmark, DualRes achieves a
1.44-fold recording-level speedup and a 24.8-fold reduction in checkpoint
storage relative to a selective state-space baseline under matched hardware
and runtime conditions.
Bearing results reveal task-dependent trade-offs. These findings support
oscillatory memory as a compact approach to vibration diagnosis under limited
labelled exposure.
\end{abstract}
\section{Introduction}
\label{sec:introduction}
Vibration sensors collect abundant operational data, but labelled recordings
of representative faults remain scarce. Windows extracted from the same
recording share a component, mounting configuration and operating condition. A useful classifier must
learn from limited labelled exposure and remain effective when these factors
change. Deployment on resource-constrained edge hardware imposes limits on model storage, memory
and inference latency.

Evaluation design is therefore central to the problem. Randomly assigning
windows from one recording to training and evaluation sets can place
near-identical signals in both; even recording-disjoint splits may contain
measurements from the same physical bearing.
\citet{vieira2026realistic} document the effects of partitioning and bearing
diversity. We report the held-out unit and unique labelled duration explicitly,
and distinguish condition transfer from unseen-component generalization.

This raises two linked questions: which compact representation captures diagnostic vibration
structure, and how should its data efficiency be evaluated? Model size alone answers neither question. A compact classifier still
needs to distinguish faults under changing conditions, and its execution cost
depends on the signal processing and inference pipeline as well as its parameters.

\begin{figure}[!h]
\centering
\includegraphics[width=\linewidth]{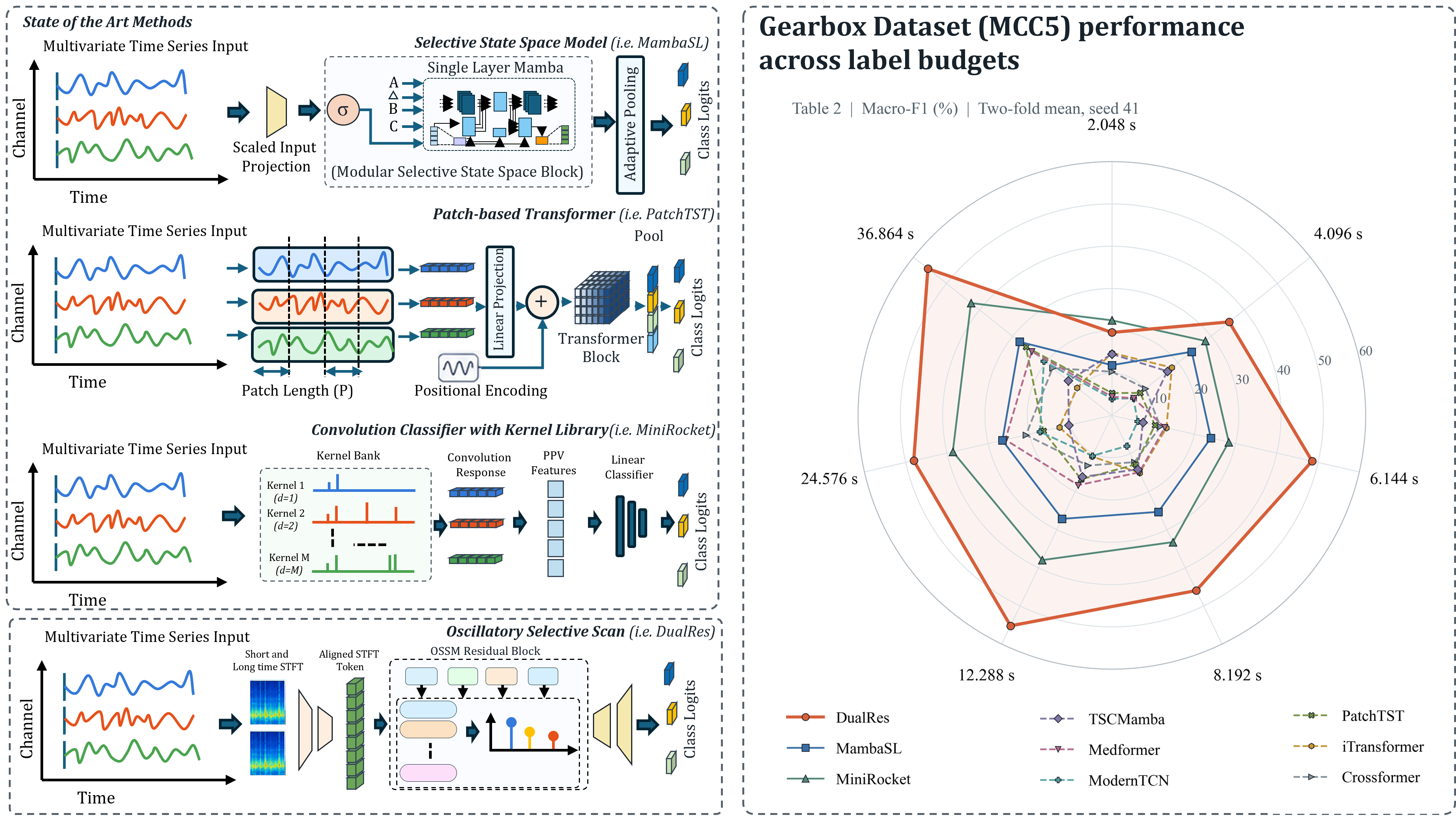}
\caption{Paper overview. Left: representative sequence classifiers and DualRes's
aligned spectral views with oscillatory memory. Right: MCC5 macro-F1 across
seven label budgets under the common supervised protocol.}
\label{fig:paper-snapshot}
\end{figure}

Mechanical vibration combines local excitations, resonant responses and temporal
modulation \citep{randall2011bearing}. Short analysis windows localize changes;
longer windows discriminate nearby frequencies. We align both resolutions at
common frame centers and model their evolution with selectively damped
oscillatory memory. We call this model DualRes, referring to its two aligned spectral resolutions. Figure~\ref{fig:paper-snapshot} summarizes
the design and its performance across labelled-data budgets.

\noindent\begin{minipage}{\linewidth}
Our contributions are threefold. \textbf{Architecture:} a center-aligned
multi-resolution classifier combines selective oscillatory memory and normalized
state updates in a $39{,}528$-parameter encoder. \textbf{Evaluation:} recording-level
splits and explicit duration accounting distinguish labelled exposure, recording
coverage and operating-condition shift. \textbf{Empirical analysis:} comparisons,
learning curves, component ablations and matched desktop measurements characterize
the accuracy--cost trade-off across bearing and gearbox tasks.
\end{minipage}

\section{Related Work}
\label{sec:related}
\paragraph{Vibration and evaluation.}
Classical bearing analysis uses recurring excitations and resonances
\citep{randall2011bearing}; public benchmarks enable data-driven evaluation
\citep{lessmeier2016paderborn,vieira2026realistic}. DualRes learns from amplitudes produced by a fixed short-time Fourier
transform (STFT), without fault-order templates or an assumption of speed invariance.
Unlike auxiliary-task meta-learning \citep{snell2017protonet,finn2017maml}, our
supervised setting restricts fitting to the stated labelled supports.

\paragraph{Sequence representations.}
Structured and complex-valued recurrent models, including S4, S4D, S5 and the
linear recurrent unit (LRU), establish compact alternatives to attention
\citep{gu2022s4,gu2022s4d,smith2023s5,orvieto2023lru}; coRNN develops coupled
oscillatory recurrence \citep{rusch2021cornn}. Mamba introduces selective updates,
and Mamba-2 relates state-space models (SSMs) to structured attention \citep{gu2023mamba,dao2024mamba2}.
MambaSL adapts a single Mamba layer for time-series classification, while TSCMamba combines wavelet and
temporal views \citep{jung2026mambasl,ahamed2024tscmamba}. Our contribution is
the specific combination of aligned spectral resolutions and normalized selective
oscillatory updates, rather than complex states or time--frequency fusion alone.

Attention models provide different ways to organize a sequence: PatchTST operates
on patches, iTransformer on variates, and Crossformer on cross-dimension structure
\citep{nie2023patchtst,liu2024itransformer,zhang2023crossformer};
Medformer and ModernTCN provide multi-granularity attention and convolutional
alternatives \citep{wang2024medformer,luo2024moderntcn}. ROCKET and MiniRocket classify sequences using convolutional feature transforms
\citep{dempster2020rocket,dempster2021minirocket}.
The comparison spans these distinct model families, following classification
benchmarking practice \citep{middlehurst2024bakeoff}.
Appendix~\ref{app:comparators} documents their classification adaptations.

\paragraph{Selection and deployment.}
Repeated evaluation can bias model selection even without within-run leakage
\citep{cawley2010selection}. Research on resource-constrained inference distinguishes parameter
counts, activation memory and measured execution cost \citep{lin2020mcunet,cai2020tinytl,banbury2021mlperf};
we therefore measure complete inference pipelines under matched conditions.

\section{Multi-Resolution Oscillatory State Space Model}
\label{sec:method}
\begin{figure}[htbp]
\centering
\includegraphics[width=\linewidth]{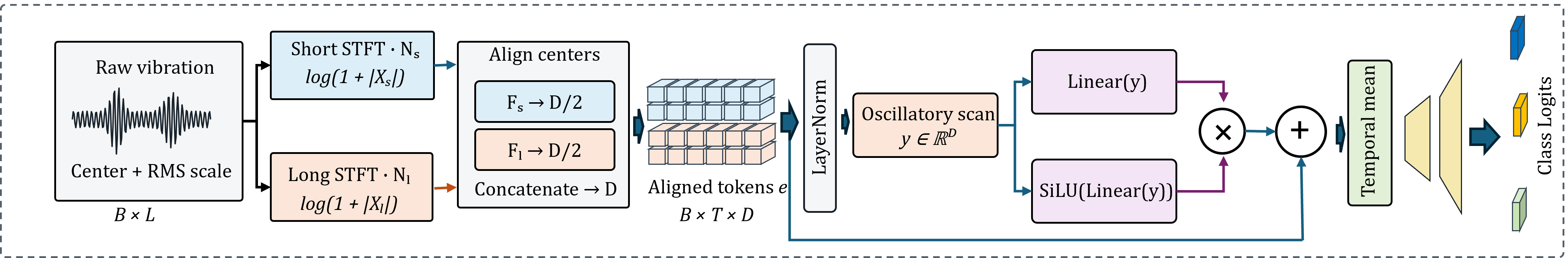}
\caption{Overall architecture: two center-aligned magnitude STFTs feed learned
projections, an oscillatory selective scan, a gated residual readout and temporal
mean pooling. The classifier is trained directly on labelled vibration.}
\label{fig:method-overview}
\end{figure}

DualRes separates local spectral analysis from temporal memory
(Figure~\ref{fig:method-overview}). It uses vibration and the acquisition sampling
rate, with no operating-condition inputs. Appendix~\ref{app:math-details}
provides full parameterizations, initialization and implementation details.

\subsection{Aligned spectral representation}
The frontend represents the same signal at two resolutions: one localizes
short-lived changes, while the other distinguishes more closely spaced
frequencies. Aligning their frame centers lets the memory process both views
as one observation at each time step.
A window $x\in\mathbb R^L$ is resampled to $f_s=64$\,kHz, centered and divided
by its root-mean-square (RMS) amplitude, using $L=32768$ ($0.512$\,s).
Normalization removes absolute amplitude while retaining within-window variation.
Let $X^{(N)}_{j,:}$ be its one-sided, unnormalized STFT with periodic Hann window,
no padding and hop $H=128$. We use $N_s=256$ and $N_l=1024$, providing
$4$ and $16$\,ms supports with $250$ and $62.5$\,Hz bin spacing
\citep{harris1978windows}.

Cropping $\kappa=(N_l-N_s)/(2H)=3$ frames from each end of the short
sequence aligns its centers with those of the long sequence:
\begin{equation}
 (j+\kappa)H+N_s/2=jH+N_l/2=\tau_j,\qquad 0\leq j<T.
 \label{eq:alignment}
\end{equation}
This gives $T=249$ aligned pairs at token rate $f_z=f_s/H=500$\,Hz.
Define $s_j=\log(1+|X^{(N_s)}_{j+\kappa,:}|)$ and
$l_j=\log(1+|X^{(N_l)}_{j,:}|)$. Affine projections form
\begin{equation}
 e_j=[W_s s_j+a_s\;\Vert\;W_l l_j+a_l]\in\mathbb R^{D},\qquad D=64,
 \label{eq:features}
\end{equation}
with $D/2$ channels per branch. All frequency bins are retained, but magnitudes
discard STFT phase. Figure~\ref{fig:resolution-alignment} illustrates the alignment.

\begin{figure}[!t]
\centering
\includegraphics[width=\linewidth]{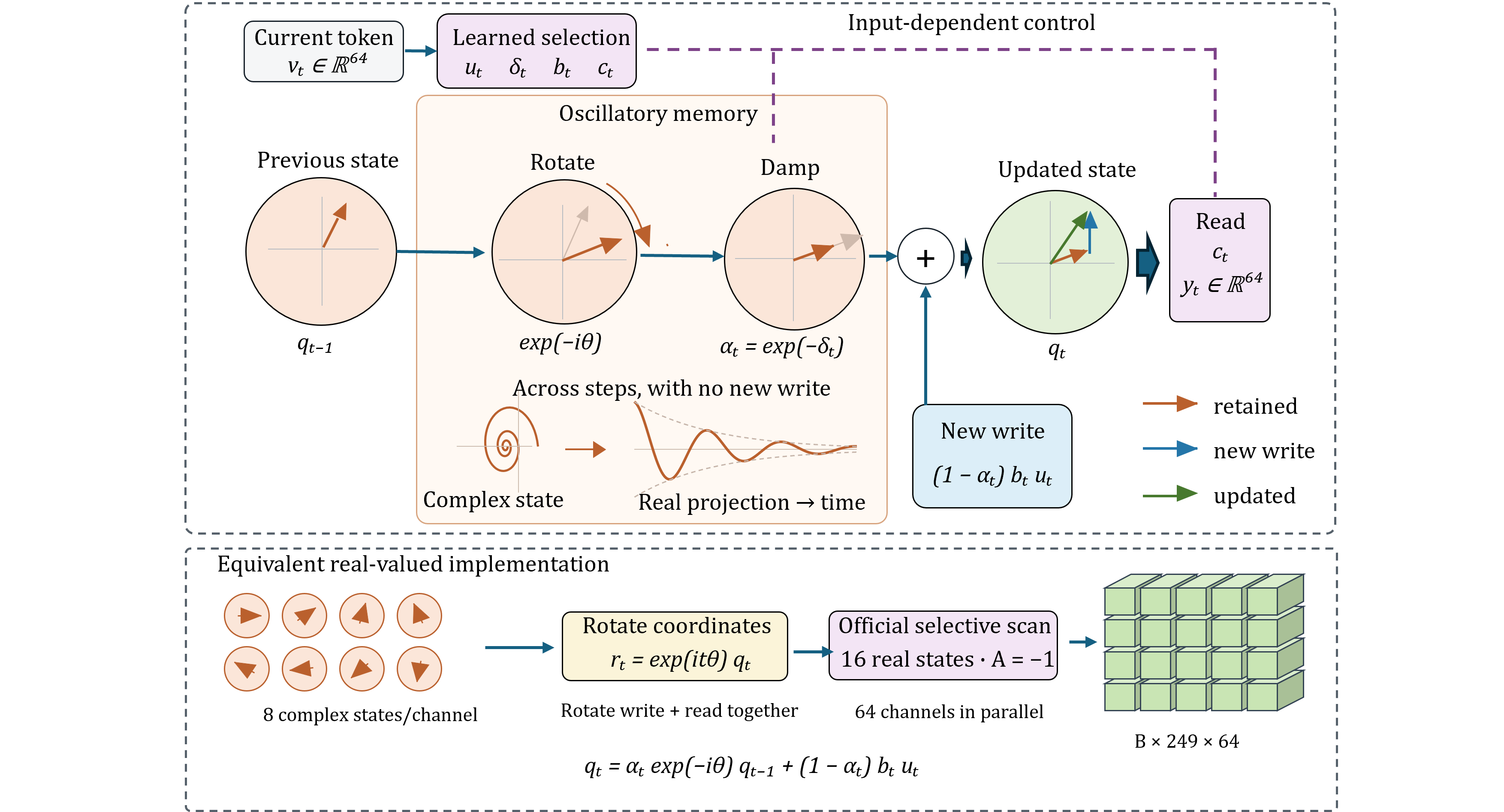}
\caption{Oscillatory memory rotates and damps the previous state, adds a
normalized write, and reads the real projection. The lower panel shows its equivalent
real-valued scan implementation; the illustrated spiral is schematic.}
\label{fig:ossm-detail}
\end{figure}

\subsection{Selective oscillatory memory}
State rotation represents oscillatory dynamics between frames, while
input-dependent damping controls memory retention. The input write is weighted
by one minus the retention coefficient, coupling new information to state decay.
Let $v_t=\operatorname{LayerNorm}(e_t)$. Learned affine maps produce input
$u_t$, positive damping increments $\delta_{t,d}$, and bounded complex write/read
vectors $b_{t,k},c_{t,k}$. Each channel $d$ maintains $K=8$ complex states,
initialized at zero:
\begin{equation}
 q_{t,d,k}=\alpha_{t,d}e^{-i\theta_k}q_{t-1,d,k}
              +(1-\alpha_{t,d})b_{t,k}u_{t,d},\qquad
 \alpha_{t,d}=e^{-\delta_{t,d}}.
 \label{eq:state}
\end{equation}
Here $\theta_k=2\pi f_k/f_z$, with learned frequencies
$f_k\in(0,f_z/2)$ shared across channels. These describe dynamics on the
\emph{spectral-token clock}, not raw carrier frequencies. The read/write vectors
are shared across channels; states and damping are channel-specific.
Figure~\ref{fig:ossm-detail} illustrates the mechanism.

As damping vanishes, the normalized write also diminishes. Bounded writes
yield bounded states by induction (Appendix~\ref{app:math-details}).

The real readout and gated residual are
\begin{align}
 y_{t,d}&=\sum_{k=1}^K\operatorname{Re}(\overline{c_{t,k}}q_{t,d,k}),\nonumber\\
 o_t&=e_t+(W_m y_t+a_m)\odot\operatorname{SiLU}(W_g y_t+a_g).
 \label{eq:output}
\end{align}
The gate transforms the memory readout; the residual preserves fused input tokens.
The official real selective-scan kernel \citep{gu2023mamba,mambasoftware}
implements Eq.~\ref{eq:state} via rotating coordinates, without an additional
standard Mamba block.

\subsection{Learning and cost}
Temporal mean pooling gives $\bar o=T^{-1}\sum_t o_t$ and logits
$\ell=W_h\operatorname{Dropout}_{0.1}(\bar o)+a_h$. All parameters are trained
jointly from labelled support $\mathcal S$ using
\begin{equation}
 \mathcal L=-|\mathcal S|^{-1}\sum_{(x,y)\in\mathcal S}
                    \log\operatorname{softmax}(\ell(x))_y.
\end{equation}
Training also applies positive multiplicative damping jitter ($\sigma=0.10$),
shared across time within each window/channel; jitter and dropout are disabled
at inference. The encoder has $39{,}528$ parameters and the head adds $65C$
for $C$ classes. The scan costs $O(TDK)$; fixed-resolution frontend and
pointwise operations also scale linearly in sequence length. Measured latency
and memory are reported separately from these analytical costs.

\section{Benchmark Design and Evaluation Protocol}
\label{sec:benchmark}
We evaluate how diagnostic performance changes with the amount and distribution
of labelled vibration. The source support is the labelled data available for
fitting; evaluation recordings are kept separate before any windows are extracted.
\paragraph{Study roadmap.}
We compare nine architectures under common label budgets, then repeat selected
comparisons across three seeds. Allocation controls separate recording diversity
from duration, and a second gearbox dataset tests the selected training recipe.
Component ablations and matched desktop profiling assess architectural
contributions and execution costs. Appendix~\ref{app:results-expanded} details
replication coverage.
\paragraph{Tasks and split units.}
We use six public bearing datasets to test changes in component identity,
operating conditions, geometry and speed profiles: Paderborn, Case Western
Reserve University (CWRU), KAIST, UORED-VAFCLS (abbreviated UORED below), HUST and Ottawa
\citep{lessmeier2016paderborn,cwru2026data,jung2023kaist,sehri2023uored,thuan2023hust,huang2019ottawa}.
The eight-class MCC5-THU gearbox benchmark (MCC5 below) varies motion
mode and fault severity \citep{chen2024mcc5}. A separate two-model pilot uses the
labelled release of the 2009 Prognostics and Health Management gearbox challenge
(PHM2009) for 14-way configuration recognition \citep{phm2009data,phm2009apparatus}.
Paderborn/UORED hold out bearing
identities; CWRU/KAIST hold out conditions; HUST holds out geometries; Ottawa
holds out speed profiles. MCC5 has two directions: low-severity torque circulation
to speed circulation (M1), and high-severity speed circulation to torque
circulation (M2). Target pools include available severities, so motion and
severity both vary. The dataset inventory, manifests and detailed split rules are in
Appendices~\ref{app:dataset-cards} and~\ref{app:protocol-details}.

\paragraph{Label budgets.}
MCC5 uses $r\in\{1,2,3,4,6\}$ recordings/class and four non-overlapping
$0.512$\,s windows/record: $2.048$--$12.288$ labelled seconds/class.
The primary budget is $3\times4$ windows ($6.144$\,s/class). A matched-duration
pilot compares $1\times12$, $3\times4$ and $6\times2$ to study recording diversity.
The extension retains six recordings and increases windows/record to eight or
twelve ($24.576$/$36.864$\,s/class). These nested supports separate the fixed-pool
duration study from joint diversity/exposure scaling. The bearing protocol uses
three overlapping windows covering $1.024$ unique seconds/class.

PHM2009 reuses the primary support budget and training duration;
Appendix~\ref{app:dataset-cards} specifies its repeat-separated folds.
\paragraph{Fitting and evaluation.}
Methods receive identical supports and vibration channels within each comparison.
All models are fitted directly on labelled source support without pretraining.
Neural MCC5/bearing runs use $300$/$100$ Adam updates, respectively; classical
methods retain their own fitting procedures. Appendix~\ref{app:training-rules}
gives optimizer settings and gradient accumulation details.
More support increases computation even at fixed update count. Terminal checkpoints
avoid target-based early stopping; repeated development on these datasets still
limits claims of independent confirmation \citep{cawley2010selection}.
The primary metric is macro-F1, the unweighted mean of per-class F1 scores;
accuracy is secondary. Both use recording-level predictions and are averaged
equally across folds. Neural models average window probabilities; MiniRocket
averages decision scores.

\paragraph{Comparisons and costs.}
Appendix~\ref{app:comparators} records classification adaptations of the released implementations. Efficiency reporting separates parameters, storage,
training time, memory and inference latency; Appendix~\ref{app:desktop} defines
the measured workloads and hardware scope.

\section{Results and Analysis}
\label{sec:results}
\subsection{Predictive performance under limited labels}
\paragraph{Bearing diagnosis.}
At 1.024 labelled seconds/class, DualRes exceeds MambaSL's mean macro-F1 on
Paderborn, KAIST and UORED, ties on Ottawa, and trails on CWRU and HUST
(Table~\ref{tab:bearing-macro_f1}). The largest positive difference is
7.71 percentage points on KAIST. UORED improves by 6.24 points but exhibits
substantial seed variability. HUST remains difficult:
MiniRocket performs best under its geometry shift.
\begin{table}[!htbp]\centering\footnotesize
\caption{Bearing macro-F1 (\%). Mean $\pm$ sample SD over three seeds, each averaged over both frozen folds. Labelled exposure: 1.024 unique seconds/class; 100 supervised updates. Bold: best three-seed mean within each dataset.}
\label{tab:bearing-macro_f1}
\setlength{\tabcolsep}{4pt}\renewcommand{\arraystretch}{1.12}
\begin{tabular*}{\linewidth}{@{\extracolsep{\fill}}lrrr}
\toprule
Dataset & DualRes & MambaSL & MiniRocket \\
\midrule
Paderborn & \textbf{48.92 $\pm$ 1.87} & 45.69 $\pm$ 2.44 & 33.50 $\pm$ 0.13 \\
CWRU & 68.04 $\pm$ 4.06 & \textbf{70.04 $\pm$ 3.03} & 66.37 $\pm$ 0.00 \\
KAIST & \textbf{45.00 $\pm$ 2.61} & 37.29 $\pm$ 3.05 & 43.37 $\pm$ 1.59 \\
UORED & \textbf{64.51 $\pm$ 15.73} & 58.27 $\pm$ 3.71 & 36.33 $\pm$ 0.00 \\
HUST & 14.16 $\pm$ 2.49 & 20.63 $\pm$ 5.71 & \textbf{22.76 $\pm$ 0.77} \\
Ottawa & \textbf{99.44 $\pm$ 0.97} & \textbf{99.44 $\pm$ 0.97} & 98.32 $\pm$ 0.00 \\
\bottomrule\end{tabular*}
\par\smallskip\begin{minipage}{\linewidth}\footnotesize iTransformer is additionally replicated on HUST (Table~\ref{tab:hust-extra-seeds}); other iTransformer datasets have one seed and are not included in this three-seed table.\end{minipage}
\end{table}

\paragraph{Gearbox diagnosis across motion modes.}
DualRes leads the nine-method MCC5 comparison at six of seven label budgets;
MiniRocket leads at 2.048 seconds/class (Table~\ref{tab:mcc5-macro_f1}).
In the replicated comparison, DualRes exceeds the stronger of MambaSL and
MiniRocket by 16.1 points at the primary budget and 11.9 at the largest.
Appendix~\ref{app:results-expanded} gives accuracy, per-fold results and repetitions.
\begin{table}[!htbp]\centering\footnotesize
\caption{MCC5 macro-F1 (\%): two-fold means, seed 41. Column headers are unique labelled seconds/class. Neural methods use 300 terminal updates; MiniRocket uses its source-fitted ridge pipeline. Bold: best completed value per budget.}
\label{tab:mcc5-macro_f1}
\setlength{\tabcolsep}{4pt}\renewcommand{\arraystretch}{1.12}
\begin{tabular*}{\linewidth}{@{\extracolsep{\fill}}lrrrrrrr}
\toprule
Method & 2.048 & 4.096 & 6.144 & 8.192 & 12.288 & 24.576 & 36.864 \\
\midrule
\textbf{DualRes} & 19.61 & \textbf{35.46} & \textbf{48.55} & \textbf{45.96} & \textbf{55.22} & \textbf{48.04} & \textbf{55.63} \\
MambaSL & 11.88 & 24.14 & 24.00 & 25.31 & 27.15 & 26.54 & 27.92 \\
MiniRocket & \textbf{22.48} & 28.24 & 28.35 & 33.25 & 37.97 & 38.57 & 42.61 \\
TSCMamba & 14.50 & 16.76 & 7.52 & 14.15 & 16.13 & 10.47 & 13.16 \\
Medformer & 4.42 & 6.54 & 12.83 & 14.94 & 18.31 & 25.83 & 24.25 \\
ModernTCN & 3.92 & 6.50 & 6.25 & 8.04 & 10.62 & 17.43 & 20.55 \\
PatchTST & 5.28 & 8.54 & 10.45 & 12.88 & 16.76 & 16.60 & 26.02 \\
iTransformer & 14.79 & 18.13 & 13.23 & 15.11 & 10.76 & 12.73 & 10.50 \\
Crossformer & 10.34 & 10.15 & 11.75 & 12.27 & 13.15 & 20.83 & 17.94 \\
\bottomrule\end{tabular*}
\par\smallskip\begin{minipage}{\linewidth}\footnotesize First five columns: 1/2/3/4/6 recordings $\times$ 4 windows. Last two: 6 recordings $\times$ 8/12 windows.\end{minipage}
\end{table}

\paragraph{A separate gearbox pilot.}
The unchanged architecture and primary training recipe yield a 24.91-point
macro-F1 advantage over MambaSL on PHM2009, with gains in both folds
(Table~\ref{tab:phm-pilot}).
\begin{table}[htbp]\centering\footnotesize
\caption{PHM2009 configuration-recognition pilot (\%). Seed 41, 3 records $\times$ 4 disjoint windows/class, 300 updates; 140 evaluation recordings per fold. Same architecture and recipe; no PHM model selection.}
\label{tab:phm-pilot}
\setlength{\tabcolsep}{3.5pt}
\begin{tabular*}{\linewidth}{@{\extracolsep{\fill}}lrrrr}
\toprule
Method & P1 F1 & P2 F1 & Mean F1 & Mean accuracy\\
\midrule
DualRes & \textbf{69.34} & \textbf{52.94} & \textbf{61.14} & \textbf{60.36}\\
MambaSL & 37.68 & 34.78 & 36.23 & 39.64\\
\bottomrule\end{tabular*}\end{table}

\subsection{How should a limited label budget be allocated?}
\begin{figure}[!htbp]\centering
\includegraphics[width=\linewidth]{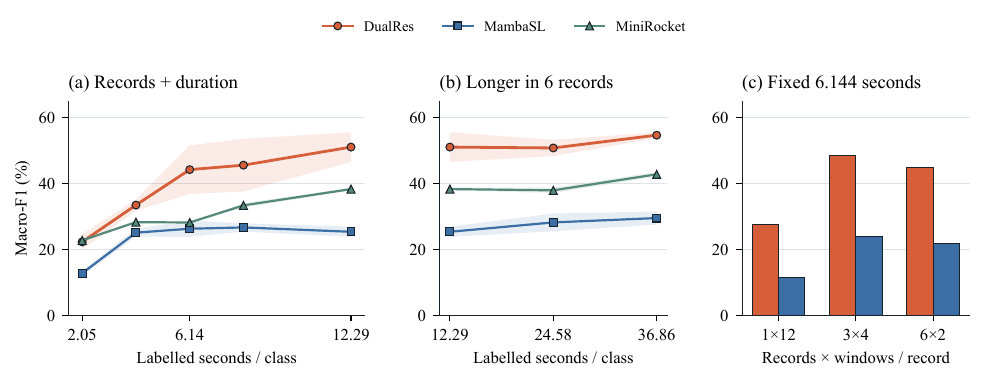}
\caption{Three different data-efficiency questions on MCC5.
(a) Increase recordings and duration jointly. (b) Increase duration in the
same six-record pool. (c) Redistribute the same 12 windows across recordings.
Panels (a,b): markers show three-seed means and shaded bands show one sample SD;
Panel (c): seed-41 bars. Every value averages both folds.}
\label{fig:results-data}
\end{figure}
DualRes improves from 22.29\% to 51.12\% mean macro-F1 as
coverage grows from one to six recordings/class (Figure~\ref{fig:results-data}a).
This changes duration, diversity and computation together. At fixed six-record
coverage, doubling duration gives little improvement, while the largest budget
reaches 54.74\%. More exposure is useful, but its effect is not monotonic.

The fixed-exposure control separates allocation from quantity. Both methods peak at
the $3\times4$ allocation (Figure~\ref{fig:results-data}c). Thus balanced within-record coverage and
between-record diversity outperform either extreme in this control.

\subsection{What do the components contribute?}
\begin{table}[htbp]\centering\footnotesize
\caption{Component ablations: recording macro-F1 (\%), seed 41, both folds. Bearing support 1.024 s/class; MCC5 support 6.144 s/class. Bold: highest score in each row; all interventions are retained.}
\label{tab:completed-ablations}
\setlength{\tabcolsep}{3.5pt}
\begin{tabular*}{\linewidth}{@{\extracolsep{\fill}}lrrrrr}
\toprule
Dataset & Full & Short only & Long only & No rotation & No jitter\\
\midrule
Paderborn & 51.07 & 47.46 & 51.85 & \textbf{52.42} & 50.69\\
CWRU & \textbf{71.28} & 60.65 & 64.14 & 68.04 & \textbf{71.28}\\
KAIST & 47.12 & 34.22 & \textbf{55.67} & 43.84 & 47.12\\
UORED & \textbf{81.23} & 54.05 & 53.61 & 54.05 & \textbf{81.23}\\
HUST & \textbf{16.65} & 5.51 & 14.33 & 11.30 & 16.16\\
Ottawa & \textbf{100.00} & \textbf{100.00} & \textbf{100.00} & \textbf{100.00} & \textbf{100.00}\\
MCC5 & 48.55 & 42.84 & \textbf{51.72} & 39.82 & 50.08\\
\bottomrule\end{tabular*}\end{table}

Matched controls remove a projected resolution, oscillator rotation or damping
jitter while retaining the backbone and training recipe
(Table~\ref{tab:completed-ablations}). Removing rotation reduces MCC5 macro-F1
by 8.73 points and UORED by 27.18 points, but slightly improves Paderborn.
The full frontend improves over either single-resolution control on CWRU,
UORED and HUST; the long-only control is better on KAIST and MCC5.
Jitter has no consistent advantage. Appendix~\ref{app:new-ablations} details
the controls and their effects across datasets.

\subsection{How do the models optimize under the same label budget?}
\begin{figure}[!htbp]\centering
\includegraphics[width=\linewidth]{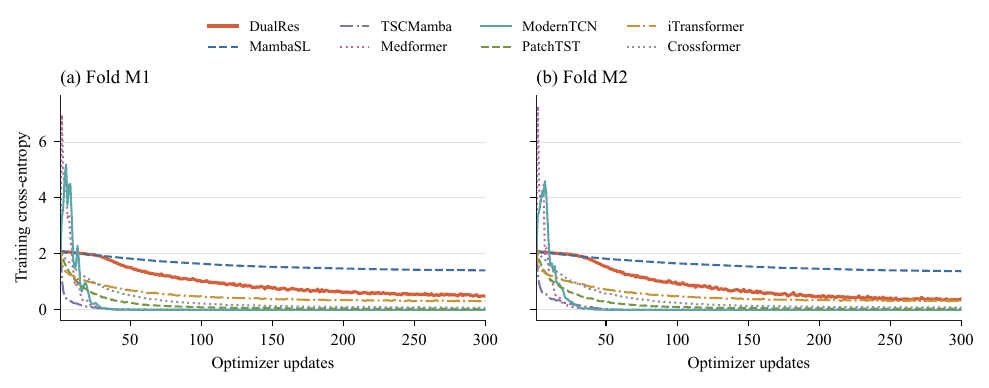}
\caption{Training cross-entropy for eight neural methods, seed 41, without
smoothing, at 12.288 labelled seconds/class (six recordings, four windows each).
Table~\ref{tab:mcc5-macro_f1} reports held-out performance.}
\label{fig:training-loss}\end{figure}
Several comparators reach near-zero training loss before DualRes yet attain
lower macro-F1 on held-out data (Figure~\ref{fig:training-loss}; Table~\ref{tab:mcc5-macro_f1}).
Lower training loss therefore does not necessarily imply better generalization.
Appendix~\ref{app:duration} documents source-only duration selection.

\subsection{Accuracy, storage and measured execution cost}
\begin{figure}[!htbp]\centering
\includegraphics[width=\linewidth]{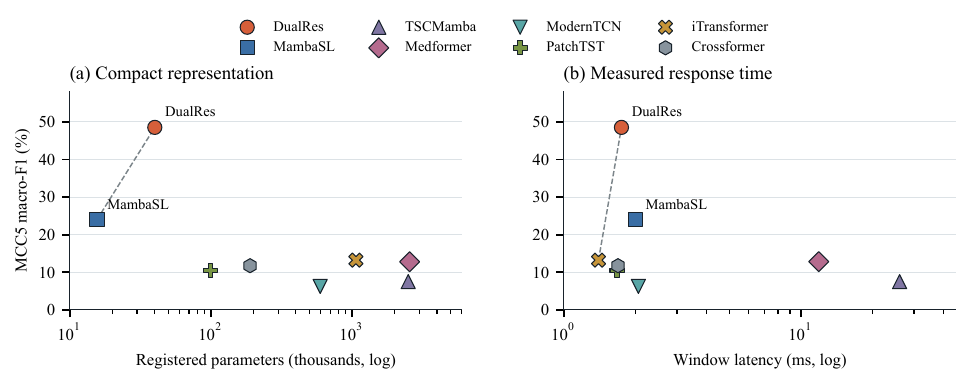}
\caption{MCC5 primary setting, seed 41: predictive quality versus
(a) registered parameters and (b) matched batch-one window latency.
Dashed segments join the nondominated observed neural models. Latency includes
model-specific preprocessing and transfers on one RTX A4000, FP32 with TF32 off.
MiniRocket's CPU costs are reported separately.}
\label{fig:results-efficiency}
\end{figure}
DualRes has 40,048 parameters including the gearbox classification head, versus
MambaSL's 15,592. Its checkpoint nevertheless requires 24.8-fold less storage
(0.164 versus 4.065\,MiB), since checkpoint size also depends on saved buffers.

On MCC5, DualRes takes 1.75\,ms/window versus 2.00\,ms for MambaSL,
and 196.65 versus 283.85\,ms per native recording (1.44-fold speedup).
Some attention models are faster but have lower macro-F1
(Figure~\ref{fig:results-efficiency}); Appendix~\ref{app:desktop} gives
all seven datasets' measurements.

\begin{table}[!htbp]\centering\footnotesize
\caption{Seven-dataset window costs: median [range] of dataset latency medians and maximum peak allocated CUDA memory. RTX A4000, batch one, FP32. Bold: lowest summary per metric. Appendix~\ref{app:desktop} gives individual datasets and CPU MiniRocket.}
\label{tab:efficiency-summary}
\setlength{\tabcolsep}{5pt}
\begin{tabular*}{\linewidth}{@{\extracolsep{\fill}}lrr}
\toprule Method & Window latency: median [range], ms & Maximum allocated MiB \\
\midrule
DualRes & 1.75 [1.56--2.63] & \textbf{11.46} \\
MambaSL & 1.65 [1.53--2.00] & 40.94 \\
TSCMamba & 22.88 [22.33--26.08] & 28.38 \\
Medformer & 11.89 [11.14--13.97] & 27.45 \\
ModernTCN & 2.06 [1.76--3.23] & 11.92 \\
PatchTST & 1.82 [1.68--2.01] & 13.92 \\
iTransformer & \textbf{1.33} [1.23--2.22] & 13.45 \\
Crossformer & 1.77 [1.64--2.26] & 14.19 \\
\bottomrule\end{tabular*}\end{table}

Table~\ref{tab:efficiency-summary} summarizes costs across all seven datasets.

\section{Conclusion and Future Work}
We presented DualRes, a compact vibration classifier that combines aligned
multi-resolution spectra with selective oscillatory memory. The two spectral
views retain complementary temporal and frequency structure, while learned
rotation and damping model their evolution in a 39,528-parameter encoder.
Across six bearing datasets and the main gearbox benchmark, the results
characterize the accuracy--cost trade-offs of learning from limited labelled
recordings. DualRes leads the nine-method gearbox comparison at six of seven
exposure budgets; bearing performance varies with the task and evaluation split.
The accompanying benchmark combines recording-level separation, explicit
labelled-duration accounting, controlled support allocation and matched
inference measurements. These studies examine both the amount of labelled data and its allocation
across recording conditions, alongside the execution costs of each model.

Future work will investigate adaptive spectral resolution for changing signal
characteristics and streaming oscillatory inference for continuous monitoring.
Additional machine types and independently collected operating conditions
will test how broadly the learned representations generalize.\label{endmain}
\clearpage
\section*{Reproducibility, Ethics and AI-Use Statements}
Appendix~\ref{app:reproduction} documents the evidence artifacts and reproduction entry points. Dataset redistribution remains governed by each source's terms. The experiments use machine signals rather than personal data; deployment would require application-specific failure analysis. AI tools assisted code development, analysis and manuscript preparation. The authors are responsible for verifying the implementation, numerical evidence, citations and final claims.
\bibliographystyle{iclr2027_conference}
\bibliography{references,draft_references}

\clearpage\appendix
\raggedbottom
\renewcommand{\topfraction}{0.9}
\renewcommand{\bottomfraction}{0.8}
\renewcommand{\textfraction}{0.08}
\renewcommand{\floatpagefraction}{0.8}
\setcounter{topnumber}{3}
\setcounter{bottomnumber}{3}
\setcounter{totalnumber}{5}
\begin{center}\Large\bfseries Supplementary Material\end{center}
\section{Mathematical and Implementation Details}
\label{app:math-details}
\subsection{Tensor shapes and parameter accounting}
\begin{table}[!htbp]\centering\small
\caption{Per-window tensor shapes; batch axis omitted.}
\begin{tabular}{lll}\toprule Stage & Shape & Parameters \\\midrule
Raw window & $32768$ & 0\\
Aligned magnitudes & $249\times129$, $249\times513$ & 0\\
Affine spectral projections & $249\times(32+32)$ & 20,608\\
LayerNorm & $249\times64$ & 128\\
Input and damping projections & two $64\to64$ maps & 8,320\\
Complex write/read projections & two $64\to16$ maps & 2,080\\
Damping rates and frequencies & $64+8$ & 72\\
Output mixing and gate & two $64\to64$ maps & 8,320\\
Temporal mean & $64$ & 0\\
Classifier & $64\to C$ & $65C$\\\midrule
Encoder total & & 39,528\\\bottomrule
\end{tabular}\end{table}
Eight complex states correspond to 16 real coordinates/channel. Shared
read/write maps account for 2,080 parameters; they are not independent
$64\times8$ projections for every channel.
\subsection{Normalization and complete spectral construction}
Let $x\in\mathbb{R}^{L}$ be a univariate vibration window after resampling to
$f_s=64{,}000$\,Hz. We use $L=32{,}768$, corresponding to $0.512$\,s.
Each window is centered and scaled by its own root-mean-square amplitude:
\begin{equation}
 \widetilde{x}_n=\frac{x_n-\mu_x}{\max(r_x,10^{-10})},\qquad
 \mu_x=\frac{1}{L}\sum_n x_n,\quad
 r_x=\sqrt{\frac{1}{L}\sum_n(x_n-\mu_x)^2}.
 \label{eq:normalization}
\end{equation}
For analysis length $N\in\{256,1024\}$ and common hop $H=128$, we compute
\begin{equation}
 X^{(N)}_{j,k}=\sum_{m=0}^{N-1}\widetilde{x}_{jH+m}
 w_N[m]e^{-2\pi i km/N},\qquad 0\leq k\leq N/2,
 \label{eq:stft}
\end{equation}
where $w_N[m]=\tfrac12(1-\cos(2\pi m/N))$ is a periodic Hann window.
Write $N_s=256$ and $N_l=1024$
for the short and long analysis lengths, and $F_s=N_s/2+1$ and $F_l=N_l/2+1$
for their one-sided frequency-bin counts. The diagrams show batch size $B$
explicitly; the equations omit that independent batch dimension.

\begin{figure}[t]
\centering
\includegraphics[width=.98\linewidth]{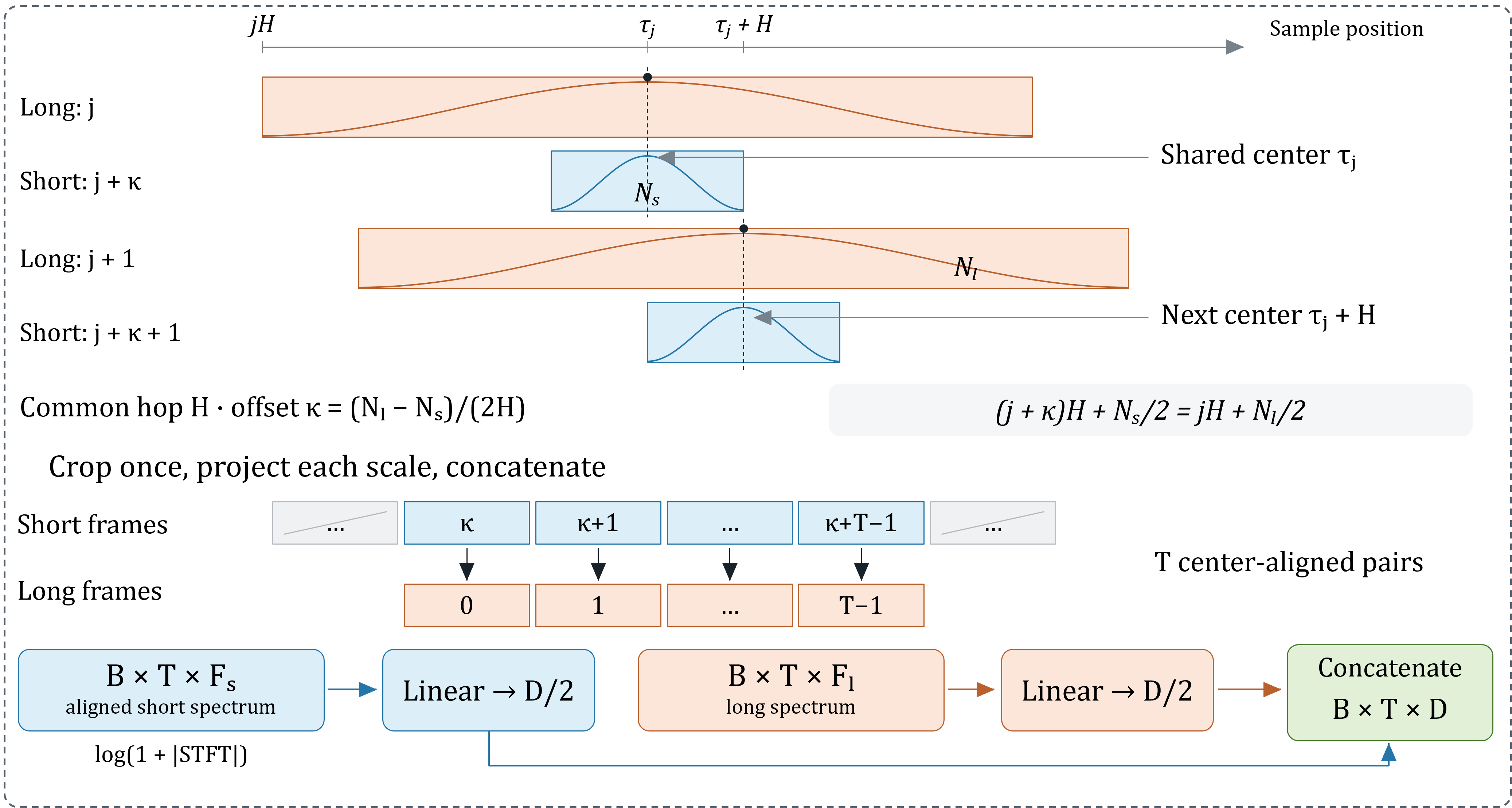}
\caption{Dual-resolution alignment. Short and long Hann windows have different
supports but paired frames share the center $\tau_j$. Cropping the short sequence
by $\kappa=(N_l-N_s)/(2H)$ gives $T$ aligned frames. Independent projections map
the $F_s$ and $F_l$ magnitude bins into $D/2$ channels before concatenation;
$B$ denotes batch size.}
\label{fig:resolution-alignment}
\end{figure}

The short transform yields $255$ frames and the long transform $249$.
Define the frame offset $\kappa=(N_l-N_s)/(2H)=3$ and the shared center
$\tau_j=jH+N_l/2$. Retaining short frames $j+\kappa$ gives the shared center
in Eq.~\ref{eq:alignment}.
Here $T=\lfloor(L-N_l)/H\rfloor+1=249$. Exact alignment by an integer-index
crop requires integral $\kappa$, as satisfied by our settings.
The resulting token rate is $f_z=f_s/H=500$\,Hz.
Figure~\ref{fig:resolution-alignment} illustrates this alignment and the subsequent
branch projections.

\subsection{Coefficient parameterization and initialization}
For token $t$, let $v_t=\operatorname{LayerNorm}(e_t)$. An affine map produces
$u_t\in\mathbb{R}^{D}$. Each channel $d$ maintains $K=8$ complex states
$q_{t,d,k}$, initialized at zero. Frequencies are shared across channels:
\begin{align}
 f_k&=\frac{f_z}{2}\operatorname{sigmoid}(\rho_k),
 &\theta_k&=2\pi f_k/f_z,\\
 \delta_{t,d}&=\frac{\exp(\lambda_d+
       \tanh([W_\gamma v_t+a_\gamma]_d))}{f_z},
 &\alpha_{t,d}&=\exp(-\delta_{t,d}).
 \label{eq:damping}
\end{align}
On the spectral-token clock defined in Section~\ref{sec:method}, initial frequencies are
log-spaced from $10$ to $200$\,Hz. Baseline damping rates $\exp(\lambda_d)$ are
initialized from half-lives log-spaced between $1$ and $64$\,ms, and the
input-dependent damping projection starts at zero.

Separate affine maps from $v_t$ to $2K$ real coordinates parameterize
\begin{align}
 b_{t,k}&=\tanh(\beta^{\mathrm{R}}_{t,k})
                +i\tanh(\beta^{\mathrm{I}}_{t,k}),\\
 c_{t,k}&=\big(\tanh(\chi^{\mathrm{R}}_{t,k})
                +i\tanh(\chi^{\mathrm{I}}_{t,k})\big)/\sqrt{K}.
\end{align}

\subsection{Bounded-state argument}
For bounded writes $|b_{t,k}u_{t,d}|\leq M$, the triangle inequality gives
\begin{equation}
 |q_{t,d,k}|\leq\alpha_{t,d}|q_{t-1,d,k}|+(1-\alpha_{t,d})M
 \leq\max\{|q_{0,d,k}|,M\}.
 \label{eq:bound}
\end{equation}
The second inequality follows by induction. This state bound assumes bounded
writes; it does not constrain every learned projection or imply generalization.
Equation~\ref{eq:state} defines a discrete update rather than an exact
discretization of a general forced continuous oscillator.

\subsection{Exact damping regularization and cost}
During training only, damping increments receive multiplicative jitter:
\begin{equation}
 \delta'_{t,d}=\delta_{t,d}\exp(\epsilon_d-\sigma^2/2),
 \qquad\epsilon_d\sim\mathcal{N}(0,\sigma^2),\quad\sigma=0.10.
\end{equation}
Noise is independent across windows and channels but shared across token positions
within a window. Its positive multiplier has mean one and preserves positive
damping; it does not preserve the expected decay coefficient itself. The perturbed
$\delta'$ defines both decay and normalized writing.

For fixed resolutions, STFT cost is
$O(T(256\log256+1024\log1024))$; projection and mixing costs are
$O(T(32(129+513)+D^2+DK))$, and the state scan costs $O(TDK)$.
All scale linearly with sequence length at fixed widths and analysis windows.
The recurrent state contains $DK$ complex values; measured activation memory
is discussed in Appendix~\ref{app:desktop}.

\subsection{Efficient implementation in rotating coordinates}
The complex recurrence can be evaluated using a real selective-scan kernel.
Factoring the known rotation out of the state leaves a real decay and moves
the phase factors into the read and write vectors.
Write $a_{t,d}=\exp(-\delta_{t,d})$ and let the recurrence be
Eq.~\ref{eq:state}. The substitution $r_{t,d,k}=e^{it\theta_k}q_{t,d,k}$ yields
\begin{equation}
 r_t=a_t r_{t-1}+(1-a_t)e^{it\theta}b_tu_t.
\end{equation}
Rotating the read vector by the same phase leaves
$\operatorname{Re}(\bar c_tq_t)$ unchanged. The official real selective scan
uses $A=-1$, $\Delta=\delta$ and a write term $\Delta Bu$. Scaling its input by
\begin{equation}
 g(\delta)=\frac{-\operatorname{expm1}(-\delta)}{\delta},
 \qquad \lim_{\delta\to0}g(\delta)=1,
\end{equation}
reproduces the normalized write without a complex CUDA kernel. This is an
algebraic representation of the chosen discrete recurrence, not a claim that
an unmodified Mamba block implements the same model.
In the homogeneous system,
$|q_t|=\exp(-\sum_{j\leq t}\delta_j)|q_0|$. Asymptotic decay requires the
cumulative damping to diverge; positivity alone does not supply a uniform
decay rate over every possible input. Neither state boundedness nor asymptotic decay ensures bounded optimization
gradients.
\subsection{Gradient paths and memory use}
Gradients pass through both spectral projections, input-dependent coefficients,
the scan, output gate and classifier. The FFT itself has no trained weights.
The full-window scan also stores sequence activations, so its memory demand
exceeds the recurrent-state footprint.

\section{Datasets and Task Definitions}
\label{app:dataset-cards}
Table~\ref{tab:task-inventory} consolidates the retained labels, acquisition rates
and evaluation split for each dataset. All tasks use one vibration channel;
metadata defines recording roles but is not an input feature. The retained label
space may be smaller than the original release.
\begin{table}[!htbp]\centering\footnotesize
\caption{Public datasets and evaluated tasks. Rates refer to the selected native
signal, before conversion to 64\,kHz. IR/OR/RE denote inner-race, outer-race and
rolling-element faults. Each task has two frozen folds.}
\label{tab:task-inventory}
\setlength{\tabcolsep}{4pt}\renewcommand{\arraystretch}{1.22}
\begin{tabularx}{\linewidth}{@{}>{\raggedright\arraybackslash}p{.16\linewidth}r>{\raggedright\arraybackslash}p{.09\linewidth}>{\raggedright\arraybackslash}X>{\raggedright\arraybackslash}p{.23\linewidth}@{}}
\toprule Dataset & Classes & kHz & Retained labels & Evaluation distinction\\\midrule
Paderborn & 3 & 64 & Healthy, IR, OR & Held-out bearing identities\\
CWRU & 4 & 48 & Healthy, IR, OR, RE & Held-out operating conditions\\
KAIST & 5 & 25.6 & Healthy, IR, OR, misalignment, unbalance & Held-out load-condition groups\\
UORED-VAFCLS & 3 & 42 & Healthy, IR, OR & Held-out bearing identities; stages grouped by bearing\\
HUST & 7 & 51.2 & Healthy, IR, OR, RE, IR+OR, IR+RE, OR+RE & Held-out bearing geometries\\
Ottawa & 5 & 200 & Healthy, IR, OR, RE, cage fault & Held-out speed-profile groups\\
MCC5-THU & 8 & 12.8 & Healthy; missing teeth; gear wear; gear pitting; tooth crack; tooth break; tooth break + IR; tooth break + OR & Opposite circulation mode; severity composition also changes\\
PHM2009 mirror & 14 & $200/3$ & Helical configurations H1--H6 and spur configurations S1--S8; case labels, not fourteen fault types & Held-out recording repeat; operating conditions also change\\
\bottomrule\end{tabularx}
\end{table}
\subsection{Bearing task boundaries}
\paragraph{Paderborn.} Combined-fault bearings are outside the retained task.
Reserved confirmation bearings \texttt{K001}, \texttt{KA30}, \texttt{KI21} and
\texttt{KB23} are rejected before signal indexing. Both artificial and real
damage occur in the release; the manifests identify the participating bearings.
\paragraph{CWRU and KAIST.} These folds test operating-condition changes, not
independent bearing populations. The fixed manifests and signal-channel choices
are common to every method.
\paragraph{UORED-VAFCLS.} All available stages of one bearing remain in one
split. Manufacturer--fault, load--fault and stage-dependent acquisition
correlations remain limitations. Centering removes constant DC offsets, not
manufacturer or gain effects. The scalar speed marker is not a tachometer
trace and is not used by the model.
\paragraph{HUST and Ottawa.} In HUST, bearing identity changes together with
geometry, so the split cannot isolate their effects. Ottawa tests speed-profile
changes on the recorded apparatus; near-perfect scores do not establish
independent-machine transfer.
\subsection{Gearbox recording roles}
\paragraph{MCC5-THU.} Version 2 contains 240 recordings. We use only
\texttt{gearbox\_vibration\_x}. Health and missing teeth each contribute 12
recordings; every other label contributes 36. M1 trains on low-severity torque
circulation and evaluates all speed-circulation recordings; M2 trains on
high-severity speed circulation and evaluates all torque-circulation recordings.
Health and missing teeth have no severity suffix. Each evaluation fold has 120
recordings; motion and severity composition change together.
\paragraph{PHM2009.} The labelled mirror contains 280 recordings: five speeds,
two loads and two repeats per case. P1 uses repeat 1 at 30/35/40\,Hz and high load
for support and evaluates repeat 2; P2 uses repeat 2 at 50/45/40\,Hz and low load
for support and evaluates repeat 1. Four disjoint windows per support recording
start at 1\,s, giving 6.144\,s/class. Both models use seed 41 and 300 terminal
updates. This expanded labelled release differs from the original unlabeled
challenge pool and its fault type/location/magnitude output
\citep{phm2009data,phm2009apparatus}. We use the first vibration column.
\subsection{Acquisition compatibility and provenance}
Native rates determine polyphase resampling to 64\,kHz (MCC5: $5/1$; PHM2009:
$24/25$). Conversion cannot restore information beyond the native Nyquist limit.
Per-window normalization follows Eq.~\ref{eq:normalization} and removes absolute
amplitude. Acquisition manifests retain original identities, release paths,
versions and available license metadata; raw redistribution remains governed
by each release's terms. Fold sizes and support accounting are specified below.

\section{Full Benchmark Specification}
\label{app:protocol-details}
\label{app:benchmark}
\subsection{Manifests, split construction and support selection}
The bearing experiments use fixed recording manifests. The
MCC5 scaling experiment uses \texttt{ROLES.json}, which enumerates the source
record indices at each level and the entire opposite-mode evaluation set. Roles
are assigned before windows are extracted. No evaluation record enters the
labelled support or the fitted classical feature scaler.
For M1 the nested source-condition order is
1000\,rpm/10\,Nm, 3000/20, 2000/10, 1000/20, 3000/10 and 2000/20.
For M2 it is 20\,Nm/3000\,rpm, 10/1000, 10/2000, 20/1000, 10/3000 and 20/2000.
The first $n$ entries are retained for every class.
\subsection{Window placement and exposure accounting}
MCC5 support windows begin at resampled index $15f_s+jL$, for
$j=0,\ldots,W-1$. Adjacent support windows are disjoint. With $n$ recordings per
class, $C=8$ classes and $L/f_s=0.512$\,s,
\begin{equation}
 N_{\mathrm{support}}=CnW,\qquad E_{\mathrm{class}}=0.512nW,
 \qquad E_{\mathrm{total}}=CE_{\mathrm{class}}.
\end{equation}
\begin{table}[!htbp]\centering\small
\caption{MCC5 exposure accounting; no overlap correction is needed for these support windows.}
\begin{tabular}{rrrrr}\toprule Records/class & Windows/record & Windows/class & s/class & Total s \\\midrule
1&4&4&2.048&16.384\\2&4&8&4.096&32.768\\3&4&12&6.144&49.152\\
4&4&16&8.192&65.536\\6&4&24&12.288&98.304\\
6&8&48&24.576&196.608\\6&12&72&36.864&294.912\\\bottomrule
\end{tabular}\end{table}
The bearing protocol uses 50\% overlap \emph{within} a labelled support
recording. For $n_w$ windows its unique duration is
$0.512+0.256(n_w-1)$ seconds/class when one support recording is used per class.
Thus three windows cover 1.024 unique seconds. Within-recording overlap does
not cross split boundaries.
\subsection{Training and checkpoint rules}
\label{app:training-rules}
Neural models use Adam with initial learning rate $5\times10^{-4}$, decay
$0.99$ per update and microbatches of four windows. Update $e$ therefore uses
learning rate $5\times10^{-4}0.99^{e-1}$. A microbatch of
$b$ windows contributes $(b/N)\mathcal L_b$ to the accumulated gradient, yielding
the support-average objective before one optimizer step. Batch-dependent layers
still see microbatch statistics; accumulation is not equivalent to a single
full-batch forward pass. Evaluation uses terminal checkpoints from completed fits.
Seed repetitions vary model initialization while keeping the development
folds and evaluation domains fixed.
\subsection{Recording-level scoring}
\label{app:scoring}
The aggregation in Section~\ref{sec:benchmark} assigns one predicted label
per recording; windows are not independent evaluation replicates.
With $C$ fixed classes, the primary metric is
\begin{equation}
 \mathrm{macroF1}=\frac1C\sum_{c=1}^{C}
 \frac{2\mathrm{TP}_c}{2\mathrm{TP}_c+\mathrm{FP}_c+\mathrm{FN}_c},
\end{equation}
with undefined class ratios set to zero. Source balance does not imply evaluation balance: MCC5 has
six health and six missing-teeth recordings per evaluation fold, versus 18 for
each other label.

\begin{figure}[!htbp]\centering
\includegraphics[width=\linewidth]{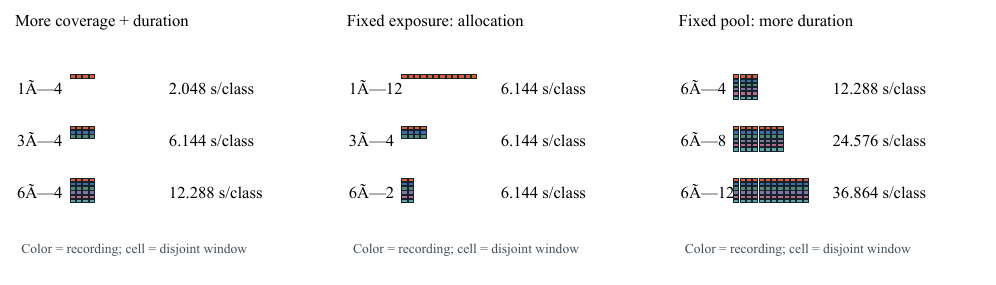}
\caption{Three support-allocation experiments. Each cell denotes a disjoint
0.512s window; each row/color within a group is a different recording.
Only source recordings supply these windows. The diagrams distinguish adding
data from reallocating a fixed duration.}

\label{fig:benchmark-protocol}
\end{figure}

\section{Comparator Configurations and Adaptations}
\label{app:comparators}
\subsection{Shared recipe and released classification paths}
All compared neural models receive the same 32768-sample support windows and
label budget under the MCC5 recipe above. The comparison is of released
classification architectures under a fixed scarce-label recipe, not a reproduction
of each paper's dataset-specific tuning grid. Internal normalization and
augmentation follow each documented implementation.
\begin{table}[!htbp]\centering\footnotesize
\caption{MCC5 comparator settings. FFN denotes the feed-forward network width. Full configurations accompany the saved fits.}
\begin{tabular}{lp{.70\linewidth}}\toprule Method & Architecture settings \\\midrule
MambaSL & Width 32, expansion 1, convolution 4; selective $\Delta,B$, fixed $C$; no $D$ skip; gating projection, 360 kernels, 8 heads.\\
PatchTST & 2 layers, width 32, FFN 128, 4 heads; patch 2048, stride 1024; released classification head.\\
iTransformer & 2 layers, width 32, FFN 64, 4 heads; released classification path on a univariate series.\\
Crossformer & 1 configured encoder layer, width 32, FFN 64, 4 heads, factor 3, segment length 1024.\\
Medformer & 6 layers, width 128, FFN 256, 8 heads; patch sizes 128/256/512; inter-granularity attention enabled; released \texttt{none,drop0.25} augmentation.\\
ModernTCN & Stage widths 32/64/128, one block/stage; kernels 9 and 5; patch 32, stride 16; dropout .5, classification dropout .1.\\
TSCMamba & One Mamba, state 128, expansion 2, convolution 4; projected space 64, image size 64, patch 8; additive fusion and max pooling; dropout .2, ROCKET disabled.\\
MiniRocket & Requested 10,000 features, actual 9,996; source-fitted centering/scaling and RidgeClassifierCV.\\
\bottomrule
\end{tabular}\end{table}
\subsection{Method-specific interpretation}
MambaSL uses a fixed released long-sequence configuration across tasks
\citep{jung2026mambasl}. TSCMamba uses the released scalar Morlet scale
\texttt{64}, with the resulting CWT resized to $64\times64$. Its raw-view orientation is repaired to match the released linear
projection, and continuous wavelet transform (CWT) extrema are fitted on source support only. The isolated
scikit-image dependency is 0.24 rather than the release's 0.22 for NumPy compatibility.
MiniRocket's source features are centered and divided by the featurewise
centered $L_2$ norm, with a unit replacement for near-zero scales. Its ridge
coefficient grid has ten log-spaced values from $10^{-3}$ to $10^3$.
MiniRocket follows its fixed-transform classifier pipeline
\citep{dempster2021minirocket}. Ridge fitting produces no iterative neural loss
curve; we report its fitted coefficients and storage separately from neural
parameter counts.
\section{Expanded Predictive Results and Replication}
\label{app:results-expanded}
\subsection{Bearing tasks and variation across folds}
\begin{table}[!htbp]\centering\footnotesize
\caption{Bearing accuracy (\%). Mean $\pm$ sample SD over three seeds, each averaged over both frozen folds. Labelled exposure: 1.024 unique seconds/class; 100 supervised updates. Bold: best three-seed mean within each dataset.}
\label{tab:bearing-accuracy}
\setlength{\tabcolsep}{4pt}\renewcommand{\arraystretch}{1.12}
\begin{tabular*}{\linewidth}{@{\extracolsep{\fill}}lrrr}
\toprule
Dataset & DualRes & MambaSL & MiniRocket \\
\midrule
Paderborn & \textbf{49.96 $\pm$ 1.13} & 47.21 $\pm$ 2.91 & 42.08 $\pm$ 0.26 \\
CWRU & \textbf{64.29 $\pm$ 3.57} & 61.90 $\pm$ 7.43 & 60.71 $\pm$ 0.00 \\
KAIST & 51.11 $\pm$ 1.92 & 40.00 $\pm$ 3.33 & \textbf{54.44 $\pm$ 1.92} \\
UORED & \textbf{64.58 $\pm$ 15.73} & 58.33 $\pm$ 3.61 & 43.75 $\pm$ 0.00 \\
HUST & 20.63 $\pm$ 1.37 & 28.17 $\pm$ 3.64 & \textbf{29.37 $\pm$ 1.37} \\
Ottawa & \textbf{99.44 $\pm$ 0.96} & \textbf{99.44 $\pm$ 0.96} & 98.33 $\pm$ 0.00 \\
\bottomrule\end{tabular*}
\par\smallskip\begin{minipage}{\linewidth}\footnotesize iTransformer is additionally replicated on HUST (Table~\ref{tab:hust-extra-seeds}); other iTransformer datasets have one seed and are not included in this three-seed table.\end{minipage}
\end{table}

\begin{figure}[!htbp]\centering
\includegraphics[width=\linewidth]{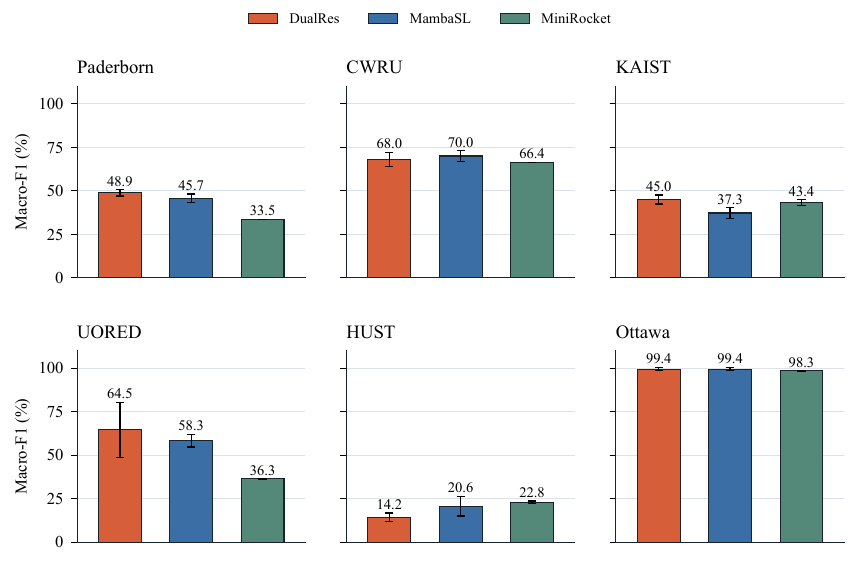}
\caption{All six bearing tasks at 1.024 unique labelled seconds/class.
Grouped bars show macro-F1 means over three fold-averaged seeds; error bars show
one sample standard deviation. All panels share the same scale. HUST
iTransformer is reported separately in Table~\ref{tab:hust-extra-seeds}.}
\end{figure}
Accuracy complements macro-F1 when evaluation classes have unequal frequencies.
\begin{table}[!htbp]\centering\footnotesize
\caption{Bearing macro-F1 (\%), seed 41, at 1.024 unique seconds/class.
Both folds use 100 supervised updates for neural models; MiniRocket fits its
ridge pipeline. Fold 1/2 follow each frozen manifest. Bold ranks the displayed
two-fold means within each dataset. This table exposes fold variation beneath
the three-seed summaries.}
\begin{tabular*}{\linewidth}{@{\extracolsep{\fill}}llrrr}\toprule
Dataset & Method & Fold 1 & Fold 2 & Mean\\\midrule
Paderborn & DualRes & 33.24 & 68.90 & \textbf{51.07} \\
 & MambaSL & 33.56 & 54.29 & 43.92 \\
 & MiniRocket & 26.70 & 40.21 & 33.46 \\
\addlinespace[3pt]
CWRU & DualRes & 69.17 & 73.40 & 71.28 \\
 & MambaSL & 76.19 & 70.00 & \textbf{73.10} \\
 & MiniRocket & 68.45 & 64.29 & 66.37 \\
\addlinespace[3pt]
KAIST & DualRes & 49.00 & 45.23 & \textbf{47.12} \\
 & MambaSL & 32.43 & 38.00 & 35.21 \\
 & MiniRocket & 57.67 & 27.23 & 42.45 \\
\addlinespace[3pt]
UORED & DualRes & 88.57 & 73.89 & \textbf{81.23} \\
 & MambaSL & 75.00 & 50.00 & 62.50 \\
 & MiniRocket & 26.19 & 46.46 & 36.33 \\
\addlinespace[3pt]
HUST & DualRes & 12.18 & 21.12 & 16.65 \\
 & MambaSL & 22.30 & 29.91 & \textbf{26.10} \\
 & MiniRocket & 23.25 & 24.02 & 23.63 \\
 & iTransformer & 22.17 & 17.07 & 19.62 \\
\addlinespace[3pt]
Ottawa & DualRes & 100.00 & 100.00 & \textbf{100.00} \\
 & MambaSL & 96.64 & 100.00 & 98.32 \\
 & MiniRocket & 96.64 & 100.00 & 98.32 \\
\bottomrule\end{tabular*}\end{table}

The seed-41 table reports the variation between folds underlying the
three-seed summary.

\subsection{Gearbox performance across label budgets}
\begin{table}[!htbp]\centering\footnotesize
\caption{MCC5 accuracy (\%): two-fold means, seed 41. Column headers are unique labelled seconds/class. Neural methods use 300 terminal updates; MiniRocket uses its source-fitted ridge pipeline. Bold: best completed value per budget.}
\label{tab:mcc5-accuracy}
\setlength{\tabcolsep}{4pt}\renewcommand{\arraystretch}{1.12}
\begin{tabular*}{\linewidth}{@{\extracolsep{\fill}}lrrrrrrr}
\toprule
Method & 2.048 & 4.096 & 6.144 & 8.192 & 12.288 & 24.576 & 36.864 \\
\midrule
\textbf{DualRes} & \textbf{24.17} & \textbf{42.92} & \textbf{50.00} & \textbf{48.33} & \textbf{59.17} & \textbf{53.33} & \textbf{58.33} \\
MambaSL & 19.17 & 27.50 & 29.17 & 31.25 & 32.92 & 32.50 & 33.33 \\
MiniRocket & 22.92 & 30.42 & 30.42 & 35.42 & 39.17 & 39.17 & 44.17 \\
TSCMamba & 17.50 & 19.17 & 13.75 & 16.67 & 17.50 & 12.50 & 15.83 \\
Medformer & 11.67 & 12.50 & 17.50 & 21.67 & 24.17 & 29.58 & 28.75 \\
ModernTCN & 10.83 & 17.50 & 14.58 & 19.17 & 18.75 & 19.17 & 21.67 \\
PatchTST & 10.42 & 18.75 & 19.17 & 21.25 & 25.42 & 25.83 & 30.00 \\
iTransformer & 15.83 & 18.75 & 14.58 & 17.08 & 11.25 & 15.83 & 12.92 \\
Crossformer & 20.83 & 19.58 & 17.08 & 18.33 & 19.17 & 25.83 & 22.50 \\
\bottomrule\end{tabular*}
\par\smallskip\begin{minipage}{\linewidth}\footnotesize First five columns: 1/2/3/4/6 recordings $\times$ 4 windows. Last two: 6 recordings $\times$ 8/12 windows.\end{minipage}
\end{table}

\begin{figure}[!htbp]\centering
\includegraphics[width=\linewidth]{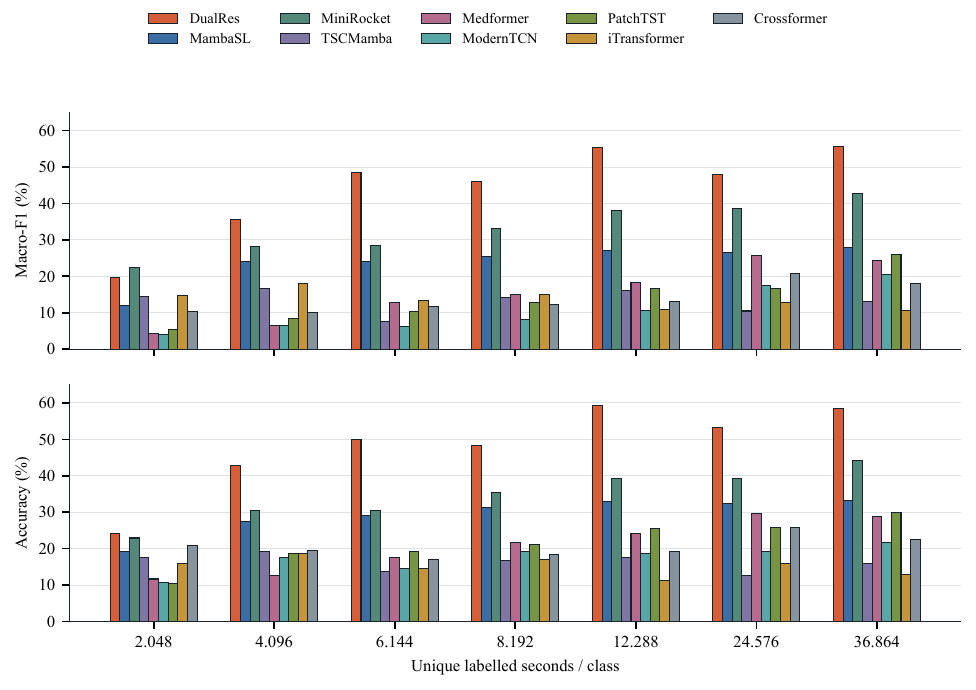}
\caption{Broad MCC5 comparison at all seven label budgets, seed 41.
Macro-F1 and accuracy are shown on a shared scale. Each bar averages both
folds. The first five
budgets increase the number of recordings, whereas
the last two increase duration within the six-record pool.}
\end{figure}
\begin{table}[!htbp]\centering\footnotesize
\caption{MCC5 macro-F1 (\%) by fold, seed41. All nine evaluated methods; bold is highest in each column.}
\label{tab:mcc5-fold-detail}
\setlength{\tabcolsep}{3.5pt}
\begin{tabular*}{\linewidth}{@{\extracolsep{\fill}}lrrrr}
\toprule
Method & 3 rec., M1 & 3 rec., M2 & 6 rec., M1 & 6 rec., M2\\
\midrule
DualRes & \textbf{42.94} & \textbf{54.16} & \textbf{46.19} & \textbf{64.26}\\
MambaSL & 22.87 & 25.13 & 20.12 & 34.19\\
MiniRocket & 17.71 & 38.99 & 23.95 & 52.00\\
TSCMamba & 8.20 & 6.83 & 13.41 & 18.85\\
Medformer & 9.44 & 16.21 & 17.58 & 19.04\\
ModernTCN & 3.31 & 9.19 & 9.38 & 11.87\\
PatchTST & 8.11 & 12.79 & 16.71 & 16.81\\
iTransformer & 9.36 & 17.10 & 8.55 & 12.97\\
Crossformer & 6.37 & 17.13 & 14.27 & 12.04\\
\bottomrule\end{tabular*}\end{table}

The per-fold table reports both directions of motion-mode transfer at the primary
and six-record settings.

\subsection{Three-seed replication coverage}
\begin{table}[!htbp]\centering\small
\caption{MCC5 replication macro-F1 (\%). Mean $\pm$ sample SD across fold-averaged seeds 41/42/43. Bold: best completed mean per row.}
\label{tab:mcc5-replication}
\setlength{\tabcolsep}{4pt}\renewcommand{\arraystretch}{1.12}
\begin{tabular*}{\linewidth}{@{\extracolsep{\fill}}lrrr}
\toprule
Seconds/class & DualRes & MambaSL & MiniRocket \\
\midrule
2.048 & 22.29 $\pm$ 2.35 & 12.78 $\pm$ 0.79 & \textbf{22.76 $\pm$ 0.24} \\
4.096 & \textbf{33.49 $\pm$ 1.72} & 25.10 $\pm$ 1.20 & 28.30 $\pm$ 0.07 \\
6.144 & \textbf{44.26 $\pm$ 7.38} & 26.30 $\pm$ 2.48 & 28.19 $\pm$ 0.15 \\
8.192 & \textbf{45.63 $\pm$ 7.99} & 26.69 $\pm$ 1.24 & 33.40 $\pm$ 0.66 \\
12.288 & \textbf{51.12 $\pm$ 4.51} & 25.36 $\pm$ 1.55 & 38.36 $\pm$ 0.41 \\
24.576 & \textbf{50.86 $\pm$ 2.48} & 28.22 $\pm$ 2.70 & 37.97 $\pm$ 0.80 \\
36.864 & \textbf{54.74 $\pm$ 0.87} & 29.54 $\pm$ 1.99 & 42.85 $\pm$ 0.72 \\
\bottomrule\end{tabular*}
\end{table}

The MCC5 replication compares DualRes, MiniRocket and MambaSL across all seven
budgets, with seeds 41, 42 and 43 and both folds per seed. Within this nine-method
scope, these are the three highest seed-41 means across the two extended budgets.
All three methods also have complete bearing repetitions. HUST additionally
replicates iTransformer, which ranks third in the broad seed-41 comparison;
Table~\ref{tab:hust-extra-seeds} reports its full seed variation. The seed-41 fit
is shared with the broad comparison at each matching configuration.
\begin{table}[!htbp]\centering\small
\caption{Additional HUST finalist replication: iTransformer at 1.024 labelled
seconds/class, 100 terminal updates. Each seed averages both frozen folds.
Its mean macro-F1 is below MiniRocket (22.76\%) and MambaSL (20.63\%), and
above DualRes (14.16\%); selection used the original seed-41 ranking.}
\label{tab:hust-extra-seeds}
\begin{tabular}{lrr}\toprule Seed & Macro-F1 (\%) & Accuracy (\%)\\\midrule
41 & 19.62 & 23.81 \\
42 & 12.39 & 16.67 \\
43 & 19.75 & 22.62 \\
Mean $\pm$ SD & 17.25 $\pm$ 4.21 & 21.03 $\pm$ 3.83\\
\bottomrule\end{tabular}\end{table}

\subsection{Generalizing the training recipe to another gearbox dataset}
\begin{figure}[!htbp]\centering
\includegraphics[width=\linewidth]{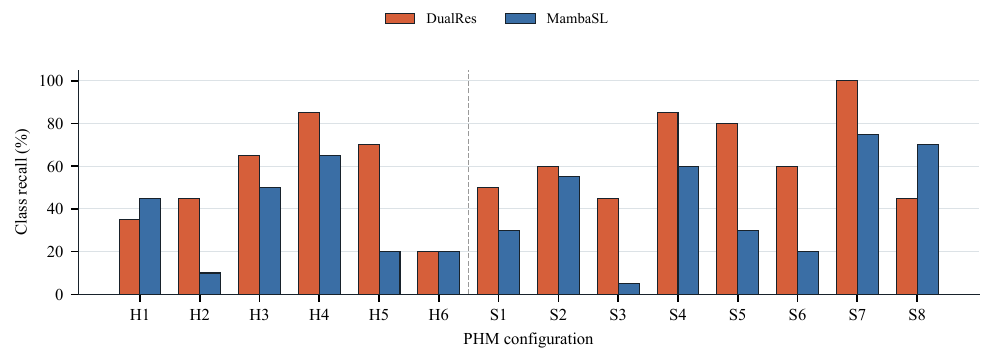}
\caption{PHM class recall, seed 41, averaged over both repeat-separated folds.
The dashed separator distinguishes helical and spur configurations. This
class-level view complements the aggregate scores in Table~\ref{tab:phm-pilot}.}
\label{fig:phm-detail}\end{figure}

All weights are fitted on PHM source support; only the MCC5 training recipe
is reused. Table~\ref{tab:phm-pilot} reports aggregate scores.
Figure~\ref{fig:phm-detail} instead identifies difficult case classes, using
recall averaged over folds. Each class has ten evaluation recordings per fold;
Table~\ref{tab:task-inventory} defines H1--H6 and S1--S8.

\section{Recording Diversity versus Labelled Exposure}
\subsection{Matched exposure control}
The allocations in Figure~\ref{fig:results-data}c share both total support
size and update count. The table below reports their scores; the recording
conditions are drawn from the same source pool.
\begin{table}[!htbp]\centering\footnotesize
\caption{Fixed 6.144 seconds/class: macro-F1 (\%), two-fold mean, seed 41.}
\begin{tabular}{lrrr}
\toprule
Records/class & Windows/record & DualRes & MambaSL \\
\midrule
1 & 12 & \textbf{27.59} & 11.44 \\
3 & 4 & \textbf{48.55} & 24.00 \\
6 & 2 & \textbf{44.82} & 21.79 \\
\bottomrule
\end{tabular}
\end{table}

The three-record allocation exceeds the one-record allocation for both methods
in both folds, but increasing from three to six recordings does not consistently improve performance.

\section{Training Dynamics and Duration Selection}
\subsection{Optimization across architectures}
Figure~\ref{fig:training-loss} in the main paper reports the cross-architecture trajectories.
Each history records the update number, support-averaged loss and learning rate.
Architecture-specific stochastic layers can affect the trajectories even
though the eight neural methods share the cross-entropy definition.
\subsection{Source-only selection of training duration}
\label{app:duration}
The separate duration pilot uses five source-mode recordings/class for training
and holds one operating condition out for validation: M1 holds
\texttt{2000rpm\_20Nm}, M2 holds \texttt{20Nm-2000rpm}.
Training uses three overlapping windows per recording (5.120 unique
seconds/class); this differs from the later non-overlapping benchmark.
Checkpoints at 50/100/200/300 updates are assessed only on these source-mode
validation files. Opposite-motion recordings are never accessed by this study.
The highest mean validation macro-F1 across both models and folds selects
300 updates, with lower log loss then earlier update as tie-breakers.
This shared duration is subsequently fixed for all MCC5 neural comparators.
\begin{figure}[!htbp]\centering
\includegraphics[width=\linewidth]{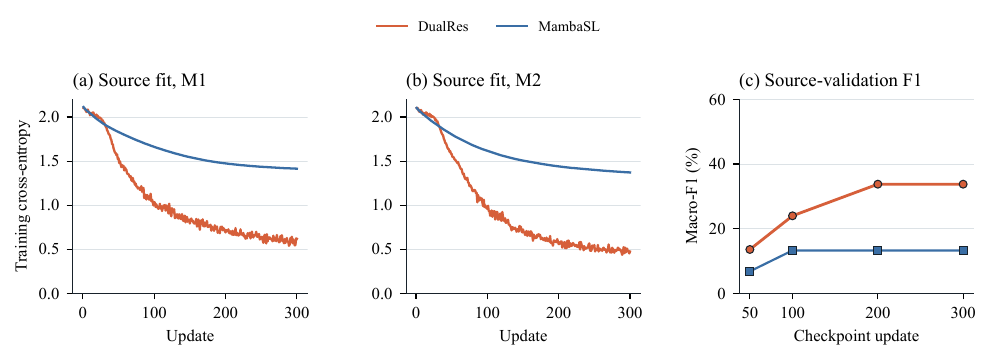}
\caption{Selecting training duration using source conditions. (a,b) Unsmoothed training losses;
(c) source-validation recording macro-F1 averaged over the two folds, seed41.
The shared 300-update choice uses both methods, not target-fold accuracy.}
\end{figure}
The histories record update indices without per-update wall-clock timestamps;
training time is therefore reported separately from these learning curves.

\section{Component Ablations}
\label{app:new-ablations}
\begin{figure}[!htbp]\centering
\includegraphics[width=\linewidth]{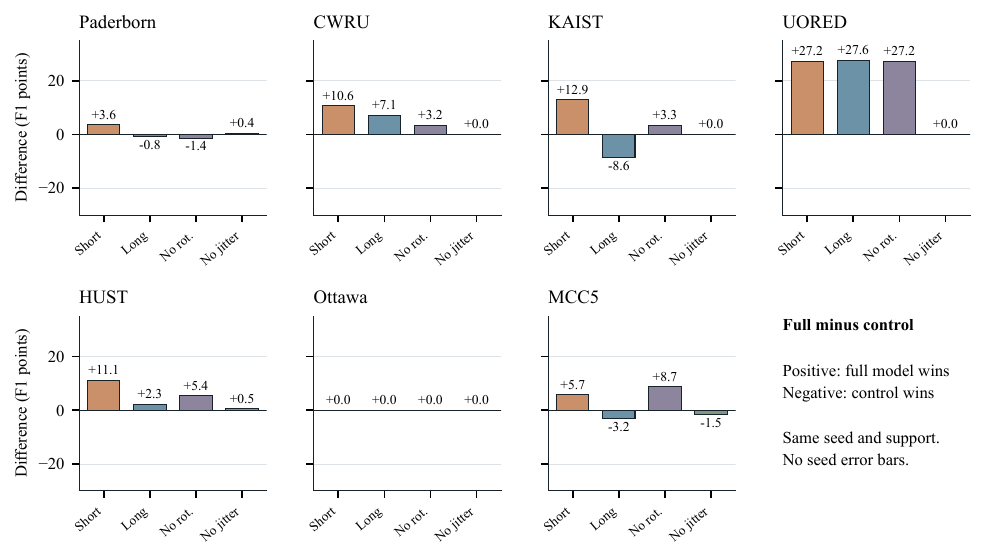}
\caption{Component contributions shown as full-model minus control macro-F1.
Positive bars favor the full model; negative bars favor the control. All seven
datasets use the same vertical scale, and zero is explicitly marked.}
\label{fig:ablation-deltas}\end{figure}

The ablation study comprises 28 control configurations and 56 fold fits.
Each is compared with the seed-41 full model at the matching support budget.
The full table is in the main paper; Figure~\ref{fig:ablation-deltas} gives signed
full-minus-control differences on every dataset. These are descriptive one-seed
effects; they do not establish uncertainty across initializations.
\subsection{Frontend resolution controls}
Short-only and long-only zero the other projected branch before LayerNorm.
Both retain the 64-channel backbone, 249 frame centers, state capacity and head.
Disabled projections are frozen, so trainable and registered counts differ.
These are controlled information removals, not optimized single-scale competitors.
\subsection{Memory and regularization controls}
The no-rotation control fixes oscillator frequencies at zero. The no-jitter
control sets jitter strength to zero while retaining the random draw,
preserving the sequence of dropout random numbers.
Pooling, initialization of retained weights and the remaining recipe stay fixed.

\section{Desktop Efficiency Measurements}
\subsection{Cost definitions and available fit measurements}
Registered parameters count all model parameters; active parameters count tensors
receiving gradients in the training path. Serialized bytes include persistent
buffers (such as positional tables) and checkpoint-format overhead, and need not
equal four bytes per parameter. GPU allocation and host-memory measurements
are defined below.
\begin{table}[!htbp]\centering\footnotesize
\caption{MCC5 training and storage costs at 6.144 labelled seconds/class (seed 41, both folds). Fit time and peak allocated training memory were recorded on a shared host; they are not isolated inference benchmarks. Bold: lowest observed neural value per column.}
\label{tab:results-costs}
\setlength{\tabcolsep}{4pt}\renewcommand{\arraystretch}{1.12}
\begin{tabular*}{\linewidth}{@{\extracolsep{\fill}}lrrrrr}
\toprule
Method & Registered & Active & File MiB & Fit s & Train MiB \\
\midrule
\textbf{DualRes} & 40,048 & 40,048 & \textbf{0.164} & \textbf{49.8} & 41.8 \\
MambaSL & \textbf{15,592} & \textbf{15,592} & 4.065 & 68.9 & 251.0 \\
MiniRocket & n/a$^\dagger$ & n/a$^\dagger$ & 0.802 & 5.2 & n/a (CPU) \\
TSCMamba & 2,502,466 & 2,502,466 & 9.555 & 100.0 & 82.2 \\
Medformer & 2,560,904 & 2,560,904 & 12.301 & 444.2 & 298.6 \\
ModernTCN & 595,112 & 595,112 & 2.308 & 65.9 & 81.2 \\
PatchTST & 99,208 & 99,208 & 4.392 & 54.7 & 38.2 \\
iTransformer & 1,066,024 & 1,066,024 & 4.079 & 53.1 & 51.6 \\
Crossformer & 189,576 & 66,440 & 4.782 & 56.0 & \textbf{37.2} \\
\bottomrule\end{tabular*}
\par\smallskip\begin{minipage}{\linewidth}\footnotesize $\dagger$ MiniRocket has 9,996 fixed-transform features and 79,976 learned linear coefficients; CPU fitting is not a neural parameter comparison. Storage includes persistent buffers and serialization overhead. GPU allocation is a framework counter, not total device memory. Matched desktop inference is reported separately.\end{minipage}
\end{table}

The training timer encloses the optimization loop, with CUDA synchronization at
its boundaries and excludes evaluation. Extended-budget evaluation recovered
from saved terminal checkpoints after CUDA errors; recovery runtime is not
reported as full-workflow training time. Training on a shared host is identified
separately from isolated inference measurements.
\subsection{Matched desktop inference}
\label{app:desktop}
We measure all 63 combinations of nine methods and seven datasets.
Each measurement uses a fresh process, four CPU/linear-algebra threads and the same
first source-support recording within its dataset. Neural paths use one RTX
A4000, FP32, TF32 disabled; MiniRocket runs on CPU and is labelled separately.
The saved checkpoint is seed41/first fold at the primary support budget.
No training or other experiment runs concurrently with these measurements.

The window workload maps a normalized CPU window to a CPU score, including
transfers and method-specific transforms. It uses 20 warm-ups and 120 timed
repetitions. Native-record timing starts from the selected channel in host memory
and includes resampling, normalization and all prescribed window predictions;
it excludes disk parsing and model loading. Both prepared-record and native-record
workloads use two warm-ups, ten repetitions and a common batch cap of four.
Bearing evaluation prefixes and the full MCC5 window count follow the original
evaluation code. Recording-level timings across datasets therefore correspond to different
workloads.

\begin{figure}[!htbp]\centering
\includegraphics[width=\linewidth]{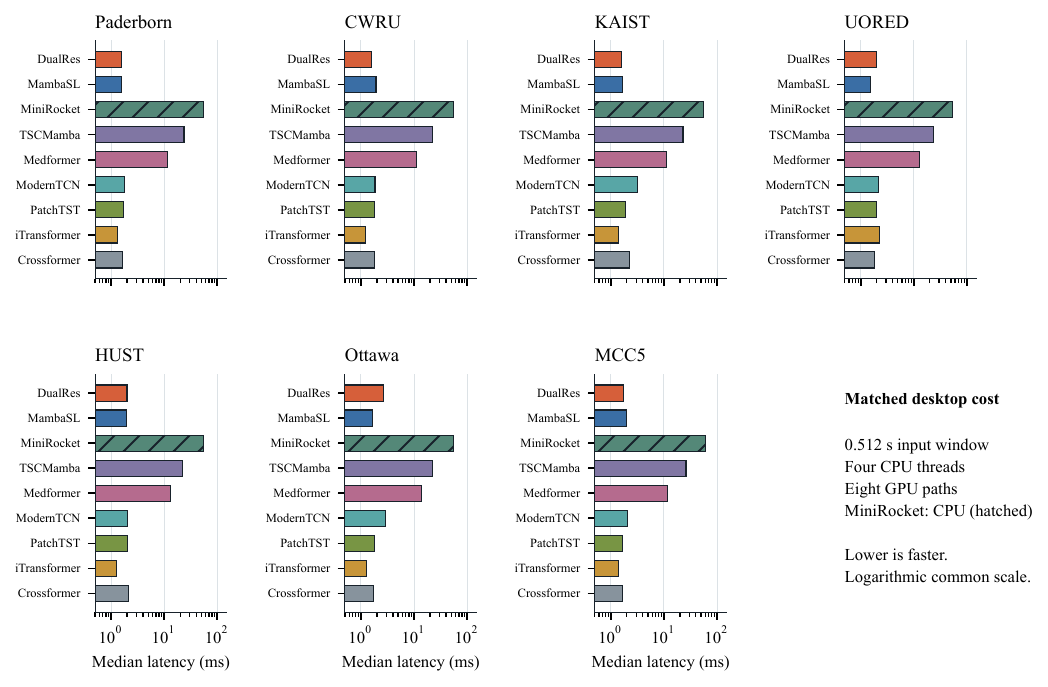}
\caption{Matched window latency for every profiled dataset and method.
Common logarithmic axes preserve large cost differences without dropping slow
methods. Hatching identifies MiniRocket's CPU backend; the other paths include
GPU inference and their required CPU work.}
\end{figure}
\begin{table}[!htbp]\centering\footnotesize
\caption{Window median latency (ms). GPU paths use one RTX A4000; MiniRocket uses four CPU threads and forms a separate backend group. Bold: lowest neural median in each dataset. Record durations differ across datasets.}
\label{tab:desktop-window}
\setlength{\tabcolsep}{3.5pt}
\begin{tabular*}{\linewidth}{@{\extracolsep{\fill}}lrrrrrrr}
\toprule
Method & Paderborn & CWRU & KAIST & UORED & HUST & Ottawa & MCC5\\
\midrule
DualRes & 1.56 & 1.61 & 1.60 & 1.96 & 1.99 & 2.63 & 1.75\\
MambaSL & 1.56 & 1.92 & 1.64 & \textbf{1.53} & 1.94 & 1.65 & 2.00\\
MiniRocket (CPU) & 55.90 & 54.52 & 54.61 & 54.20 & 54.03 & 54.48 & 60.21\\
TSCMamba & 23.63 & 22.48 & 22.88 & 23.31 & 22.33 & 22.42 & 26.08\\
Medformer & 11.35 & 11.19 & 11.14 & 13.03 & 13.23 & 13.97 & 11.89\\
ModernTCN & 1.76 & 1.84 & 3.23 & 2.15 & 2.00 & 2.95 & 2.06\\
PatchTST & 1.73 & 1.77 & 1.89 & 1.99 & 2.01 & 1.82 & 1.68\\
iTransformer & \textbf{1.33} & \textbf{1.23} & \textbf{1.37} & 2.22 & \textbf{1.24} & \textbf{1.28} & \textbf{1.40}\\
Crossformer & 1.64 & 1.77 & 2.26 & 1.84 & 2.16 & 1.75 & 1.69\\
\bottomrule\end{tabular*}\end{table}

\begin{table}[!htbp]\centering\footnotesize
\caption{Native-record median latency (ms). GPU paths use one RTX A4000; MiniRocket uses four CPU threads and forms a separate backend group. Bold: lowest neural median in each dataset. Record durations differ across datasets.}
\label{tab:desktop-native_record}
\setlength{\tabcolsep}{3.5pt}
\begin{tabular*}{\linewidth}{@{\extracolsep{\fill}}lrrrrrrr}
\toprule
Method & Paderborn & CWRU & KAIST & UORED & HUST & Ottawa & MCC5\\
\midrule
DualRes & 8.22 & 21.70 & \textbf{146.51} & \textbf{24.44} & \textbf{12.34} & 41.44 & 196.65\\
MambaSL & 12.92 & 29.05 & 152.33 & 39.19 & 14.95 & 51.25 & 283.85\\
MiniRocket (CPU) & 255.27 & 474.24 & 556.55 & 581.61 & 125.75 & 584.36 & 4049.40\\
TSCMamba & 329.44 & 505.36 & 658.93 & 762.50 & 164.65 & 771.09 & 5361.47\\
Medformer & 50.61 & 89.81 & 226.66 & 173.33 & 36.95 & 153.72 & 1069.51\\
ModernTCN & 14.27 & 23.46 & 155.97 & 26.18 & 13.32 & 40.25 & 232.70\\
PatchTST & 7.45 & 26.07 & 153.04 & 28.38 & 14.23 & 47.61 & 220.23\\
iTransformer & \textbf{6.05} & \textbf{19.53} & 150.25 & 28.26 & 18.54 & \textbf{35.40} & \textbf{184.22}\\
Crossformer & 10.12 & 21.35 & 146.97 & 30.07 & 14.46 & 36.99 & 210.69\\
\bottomrule\end{tabular*}\end{table}

\begin{table}[!htbp]\centering\footnotesize
\caption{Absolute peak allocated CUDA memory (MiB), batch-one window workload. Includes loaded tensors and temporary allocations; excludes the CUDA context and driver memory. Bold: smallest neural value per dataset.}
\label{tab:desktop-memory}
\setlength{\tabcolsep}{3.5pt}
\begin{tabular*}{\linewidth}{@{\extracolsep{\fill}}lrrrrrrr}
\toprule
Method & Paderborn & CWRU & KAIST & UORED & HUST & Ottawa & MCC5\\
\midrule
DualRes & 11.46 & 11.46 & 11.46 & 11.46 & \textbf{11.46} & 11.46 & \textbf{11.46}\\
MambaSL & 40.94 & 40.94 & 40.94 & 40.94 & 40.94 & 40.94 & 40.94\\
TSCMamba & 28.38 & 28.38 & 28.38 & 28.38 & 28.38 & 28.38 & 28.38\\
Medformer & 26.35 & 26.57 & 26.79 & 26.35 & 27.23 & 26.79 & 27.45\\
ModernTCN & \textbf{10.67} & \textbf{10.92} & \textbf{11.17} & \textbf{10.67} & 11.67 & \textbf{11.17} & 11.92\\
PatchTST & 13.90 & 13.90 & 13.90 & 13.90 & 13.91 & 13.90 & 13.92\\
iTransformer & 13.45 & 13.45 & 13.45 & 13.45 & 13.45 & 13.45 & 13.45\\
Crossformer & 14.17 & 14.18 & 14.18 & 14.17 & 14.19 & 14.18 & 14.19\\
\bottomrule\end{tabular*}\end{table}

\begin{table}[!htbp]\centering\footnotesize
\caption{Profiled source workloads and DualRes tail latency. The 95th-percentile (P95) estimates use 120 window and 10 record repetitions. RSS is the sampled absolute process resident memory and is not model-only storage.}
\label{tab:workloads}
\setlength{\tabcolsep}{3.5pt}
\begin{tabular*}{\linewidth}{@{\extracolsep{\fill}}lrrrrr}
\toprule
Dataset & Native s & Windows & Win. p95 ms & Record p95 ms & RSS MiB\\
\midrule
Paderborn & 4.00 & 14 & 2.10 & 9.80 & 853.55\\
CWRU & 10.11 & 23 & 2.12 & 26.01 & 868.90\\
KAIST & 120.00 & 23 & 2.76 & 153.05 & 989.84\\
UORED & 10.00 & 35 & 2.38 & 36.10 & 880.24\\
HUST & 10.00 & 7 & 2.86 & 13.92 & 871.34\\
Ottawa & 10.00 & 35 & 3.16 & 49.19 & 911.23\\
MCC5 & 60.00 & 233 & 2.69 & 232.98 & 1012.18\\
\bottomrule\end{tabular*}\end{table}

Memory is measured in a separate untimed pass. CUDA allocated and reserved
bytes are distinct allocator counters; neither measures the full device or
operating-system footprint. Process resident set size (RSS) is sampled every 10 ms and may miss
shorter peaks. The exported ledger also contains prepared-record timings,
incremental memory, buffers, head/encoder counts, raw timing repetitions,
input/checkpoint hashes and prediction agreement checks. Zero incremental RSS
does not mean zero memory demand: previously allocated memory can be reused.
Tail timings are descriptive estimates from these finite repetition counts.

\section{Benchmark Reproduction and Artifact Provenance}
\label{app:reproduction}
\subsection{Artifacts and implementation entry points}
The benchmark root \path{research/mcc5_all13_scaling/}
provides \texttt{run\_neural}, \texttt{run\_tscmamba} and
\texttt{run\_minirocket}; \path{research/mcc5_extended_exposure/}
contains the \texttt{windows8} and \texttt{windows12} extensions.
From the repository root:
\begin{quote}\small\ttfamily
python -m research.mcc5\_all13\_scaling.run\_neural dualres --gpu 0
\end{quote}
The model is \path{research/stft_dualres_ossm/model.py};
\path{research/dualres_seed41_reference/} is the architecture reference.
The package contains standalone DualRes code, comparator adapters, manifests and
manuscript sources. Remaining local paths require adjustment for portable use.

\subsection{Seeds, runtime and evidence provenance}
Broad MCC5 results use model seed 41 and support permutation seed 1041;
replications add model seeds 42 and 43. Encoder and classifier initializations both use the requested seed.
The runtime is WSL Ubuntu 22.04, Python 3.10, official \texttt{mamba\_ssm}
selective scan and RTX A4000 GPUs. Per-job plans record source hashes.

Result-source paths and SHA-256 hashes are recorded in
\texttt{COMPLETED\_EVIDENCE.json}, \texttt{COMPLETED\_ADDITIONS.json}
and \texttt{APPENDIX\_EVIDENCE.json}. Tables and plots use these completed results;
ablations use the matched campaign rather than architecture-search trials.

\end{document}